\documentclass{article}

\usepackage[final]{neurips_2019}
\usepackage[utf8]{inputenc} % allow utf-8 input
\usepackage[T1]{fontenc}    % use 8-bit T1 fonts
\usepackage{hyperref}       % hyperlinks
\usepackage{url}            % simple URL typesetting
\usepackage{booktabs}       % professional-quality tables
\usepackage{amsfonts}       % blackboard math symbols
\usepackage{nicefrac}       % compact symbols for 1/2, etc.
\usepackage{microtype}      % microtypography
\usepackage{graphicx}
\usepackage{paralist}

\newcommand{\eval}{\mathrm{eval}}
\newcommand{\bftuple}[2]{\langle \mathbf{#1}, \mathbf{#2} \rangle}

\title{Preventing Adversarial Use of Datasets through Fair Core-Set Construction}

\newcommand{\hide}{\mathrm{hide}}

\author{%
  Benjamin Spector\thanks{Work done as an intern in Google Research, Mountain View, CA.} \\
  MIT\\
  Cambridge, MA 02139  \\
  \texttt{spectorb@mit.edu} \\
  \And
  Ravi Kumar \\
  Google Research \\
  Mountain View, CA 94043 \\
  \texttt{ravi.k53@gmail.com} \\
  \And
  Andrew Tomkins \\
  Google Research \\
  Mountain View, CA 94043 \\
  \texttt{atomkins@gmail.com} \\
}

\begin{document}

\maketitle

\begin{abstract}
  We propose improving the privacy properties of a dataset by
  publishing only a strategically chosen "core-set" of the data
  containing a subset of the instances.  The core-set allows strong
  performance on primary tasks, but forces poor performance on
  unwanted tasks. We give methods for both linear models and neural
  networks and demonstrate their efficacy on data.
\end{abstract}

\section{Introduction}

The advent of deep learning has led the field of computer science to a difficult crossroads. On the one hand, in pursuit of performance state-of-the-art methods require ever-larger mountains of data. On the other, users have become more conscious and protective of their data. Much of this concern is over fear of what could be done with data: most are in favor of more relevant search results and recommendations, but many would be worried about malicious use of their personal data.

Most current works take approaches related to federated learning or differential privacy \citep{dwork2011differential, bonawitz2019towards, geyer2017differentially}. In this work, we take an alternative approach, most similar to \citet{schmidt2018fair}, by reducing a dataset to a much smaller subset which is most applicable to the task at hand and least applicable to other tasks.

Our framework operates as follows. First we train one or more models
for the entire dataset; we treat these models as approximations of the
target function, constrained to lie within the model family. Second,
we phrase an optimization problem to select a core-set that supports
reconstruction of the approximate target while simultaneously
optimizing for desirable secondary criteria.  In this abstract, we
focus on the secondary criterion of preserving privacy of specified
target attributes, but this applies more broadly.

The benefits of a core-set approach to privacy are threefold. First, by storing much smaller amounts of data, the potential downside of the training data being compromised is significantly lessened. (Incidentally, this also has many auxiliary advantages, such as lowered costs and footprint for data storage, and more rapid training.) Second, it is also possible to construct core-sets that increase the difficulty of solving problems other than the specific problem of interest, lessening privacy concerns. Third, it is simple and practical, because all it requires is deletion of certain data elements.

\section{Background}

Core-sets are popular in classical machine learning and data mining,
typically for increasing the speed of
algorithms. \citep{har2007smaller,tsang2005core} However, they remain
relatively unexplored for both neural networks and other applications.

Given a training set $X = \{x_1, \ldots, x_n\}$ of examples and corresponding $Y = \{y_1, \ldots, y_n\}$ of labels, we train a model $M_{X,Y}$ to predict on unseen $x_{n+1},\ldots$. We are provided a model evaluator $\eval(\cdot)$ (typically based on a testing set) that maps a model to a scalar performance. The \emph{core-set} problem is to find indices $I=\{i_1, i_2, \ldots \}$, $|I| \ll |X|$, inducing subset $X_I = \{x_{i_1}, x_{i_2}, \ldots \}$ of the examples and $Y_I = \{y_{i_1}, y_{i_2}, \ldots \}$ of the labels  that maximize $\eval(M_{X_I,Y_I})$. Analogously, we define the \emph{$k$-core-set} problem as the same problem with $|I|=k$.  Throughout the paper, $k$ will denote the core-set size.

\section{Methods}

One might reasonably wonder whether core-sets could ever perform
significantly better than random samples.  As a trivial example,
consider the problem of estimating the mean and variance of a normal
distribution $\sim N(\mu, \sigma)$ given $n$ samples. As the samples
are unbiased, our best-estimate mean should be the mean of the
samples, an unbiased estimator with expected squared error of
$\sigma^2/n$. As $n$ grows very large we will have an excellent
estimate of $\mu$. Likewise, standard techniques allow us to estimate
the variance of the underlying distribution.  Given $n$ samples, we
may for example select a core-set of $k$ elements to preserve the
sample mean, outperforming a random sample.  We may also in this
core-set carefully select points with greater dispersion than random,
obscuring the variance of the distribution.  Such a core-set would be
appropriate to publish if the goal is to allow learning of the mean,
while protecting the variance.

Thus, by condensing the greater information of the full sample into a small number of examples through careful choice, we may produce a (non-representative) sample that nonetheless achieves superior performance on our evaluation metric. Furthermore, given this great freedom, we may also select unrepresentative samples that disguise other information about the true distribution. 

What enabled us to successfully condense and disguise our dataset was our knowledge of the learning method. Our algorithm did not require the sample variance; therefore we did not sacrifice performance by altering it.  One could imagine alternate approaches that make use of the sample variance, such that obscuring it might damage performance on the task of mean estimation.  While we cannot cover this in more detail here, as in other areas of privacy, privacy-preserving core-sets require striking a balance between accuracy and privacy.

\section{A synthetic case study using linear regression}

Suppose we have a dataset $X$, target values $Y$ which we would like
to predict, and targets $Z$ which we would like to hide. We propose to
build a loss that encourages fidelity in reconstructing $Y$ while
obscuring $Z$.  Specifically, we use the following scheme for core-set
construction, following our framework. First, train on the entire
dataset, and record the weights $\mathbf{w}$ and the intercept
$w_0$. Then, define a loss for a point as $\ell(\mathbf{x}, y, z) = (y -
\bftuple{w}{x} - w_0)^2 + \alpha \ell_\hide(\mathbf{x},z)$, where $\mathbf{x}
\in X, y \in Y, z \in Z$. The first term of the loss favors
points that align with the learned model for $Y$ from the full data.
The second term, not yet specified, should make it hard to
recover $Z$.  To construct a $k$-core-set, simply choose the $k$
points with lowest loss.

For the term $\ell_\hide$, many approaches are possible, but we consider two
simple approaches:
\begin{compactitem}
    \item \emph{Hide the secret value}.  In this approach, we take $\ell_\hide(\mathbf{x},z) = (z - \mathop{\mathbb{E}}[z])^2$. This is a simple way to make $z$ difficult to predict, by choosing samples that are all roughly the same. However, this method might select a core-set that is artificially uniform in $z$ value, perhaps selecting almost all points from a majority class.
    \item \emph{Plant a synthetic secret value}.  In this approach, we take $\ell_\hide(\mathbf{x},z) = (y - \bftuple{v}{x} - v_0)^2$ for some random vector $\mathbf{v}$ and intercept $v_0$. As  the core-set is a subset of data points with their labels, we cannot modify the labels of points.  Instead, we favor selection of points whose noise in the $z$ direction will mislead a learner.  This will encourage simple learners to discover the planted value rather than the original value.
\end{compactitem}

In this short paper, we don't argue that these two simple approaches
are safe against targeted attacks; in fact, we expect they are not.
However, we can analyze their effect on standard training algorithms.
Let $\mathbf{x} = (x_1, x_2, x_3)$ with $x_1 \sim N(0,1), x_2 \sim
\mathrm{Uniform}(-1,1), x_3 \sim \mathrm{Exp}(\lambda=1)$ and labels $y \sim \bftuple{w}{x}+w_0+N(0,0.5)$ and $z \sim \bftuple{v}{x}+v_0+N(0,0.5)$, where
each $w_i,v_i \sim N(0,1)$.  The task is to recover the line of best
fit. As seen in Table 1, for a full dataset of size $1000$ and $k=50$,
our method works well, preserving prediction of $Y$ while masking the
true relationship between $X$ and $Z$.

\begin{table}
\small
\begin{tabular}{c|c|c}
\hline
Data & $y$ & $z$ \\
\hline
True values & $-0.886-1.05x_1-1.31x_2-0.586x_3$ & $-1.55+0.318x_1-0.912x_2+0.993x_3$ \\
& $+N(0,0.5)$ & $+N(0,0.5)$ \\
\hline
Full dataset & $-0.863-1.04x_1-1.28x_2-0.607x_3$ & $-1.53+0.317x_1-0.923x_2+0.982x_3$ \\
($r^2$) & ($0.8913$) & ($0.8638$) \\
\hline
Core-set & $-0.803-0.984x_1-1.39x_2-0.570x_3$ & $-2.00-0.756x_1+0.0113x_2+0.678x_3$ \\
($r^2_{\mathrm{coreset}}, r^2_{\mathrm{full}}$) & ($0.9901,0.8838$) & ($0.9755,-0.3522$) \\
\hline
\end{tabular}
\caption{\small Linear regression performance for each task, training on both the full dataset as well as the core-set. The random regression used was $z' = -2.09-0.898x_1+0.129x_2+0.651x_3$. The key result is that the full dataset is predictive for both problems, but the core-set is only predictive for $y$.}
\vspace*{-5mm}
\end{table}

\normalsize

\begin{figure}[h]
\begin{center}
\includegraphics[scale=0.7]{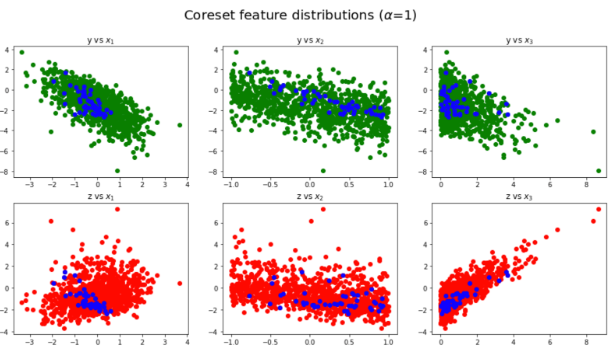}
\end{center}
\caption{\small Visualization of the datasets (green/red) and core-sets (blue) for each feature and label from Table 1.}
\end{figure}

\section{Core-sets for CIFAR-100}

The synthetic example above shows that core-sets can hide properties of the original dataset, but the example is stylized, and the $y$ and $z$ values are chosen independently.  We now consider a real dataset, and attempt to obscure one set of labels that is highly correlated with another.

CIFAR-100 comes labeled both with coarse and fine-grained labels. So, we will try to select a subset of data such that performance on identifying coarse-grained classes is maximized but performance on fine-grained classification is minimized. Additionally, to ensure that classes are not simply dropped from the dataset, we also constrain the dataset to be class-balanced under the fine-grained labels.

\begin{figure}
\begin{center}
\includegraphics[scale=0.4]{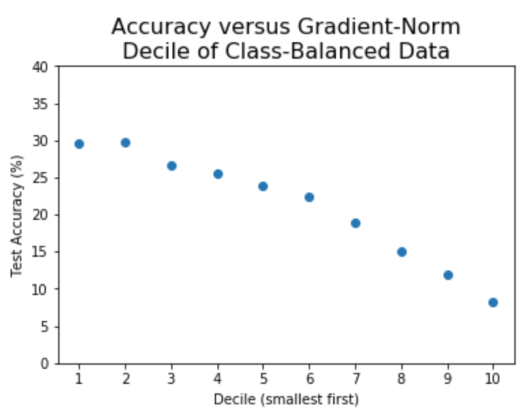}
\end{center}
\caption{\small CNN CIFAR-100 test accuracy trained on $k=5000$ class-balanced core-sets versus the average gradient norm of the examples used. Core-sets made of smaller-normed examples do better.  \label{fig:cifark}}
\end{figure}

Again following our framework, we train a model on the full dataset.  We collect for each example the \emph{average gradient magnitude} over training as a proxy for its difficulty, with the idea being that difficult, unrepresentative or uncommon examples will yield lower performance for small $k$ than easy examples. (Evidence for this method is presented in Figure~\ref{fig:cifark}.) So, we will choose examples that produce good core-sets for coarse-grained classification but generalize poorly on the individual classes.

We trained a wide-resnet-16-8 on CIFAR-100, and calculated the average gradient norm of each example on each task. We then constructed several core-sets for evaluation.  The \emph{fine-grained-masking core-set} is constructed by sorting and class-balancing data by the quotient of an example's average fine-grained gradient norm and its coarse-grained norm, and the \emph{coarse-grained-masking core-set} by the inverse of that quotient. For completeness, we list results using a \emph{random core-set} and a \emph{high-quality core-set} for each task using the minimum gradient norm examples. Final accuracy were obtained using a small CNN due to compute limitations.

As seen in Table 2, we found that we were successful in constructing core-sets which would maximize performance for one task and minimize for the other.

\begin{table}[h]
\centering
\begin{tabular}{r|c|c}
\hline
Core-set & Coarse-grained & Fine-grained \\
($k=5000$) & classification (\%) & classification (\%) \\
\hline
Random & 36.35 & 22.83 \\
Fine-grained min-norm & 42.86 & 30.79 \\
Coarse-grained min-norm & 43.10 & 27.15 \\
Fine-grained-masking & 39.55 & 16.40 \\
Coarse-grained-masking & 33.53 & 26.32 \\
\hline
\end{tabular}
\caption{\small Performance of CNN classifier on various core-sets. The bottom two rows evidence that even on these related tasks our method can construct core-sets which significantly favor one task over the other.}
\vspace*{-8mm}
\end{table}

\section{Discussion}

The main implication of this work is that one need not alter or
synthesize data nor come up with alternate embeddings for data as in
\citet{zemel2013learning,wang2018dataset}.  Rather, by carefully
selecting an appropriate subset of data for release, one can
sufficiently alter the statistics of relevant properties so as to
obscure protected values.  In this extended abstract, we introduce
these ideas without discussing specifics of attacks and
counter-attacks, as the start of a conversation about core-set-based
privacy.

In the linear regression domain, our method successfully hides
protected functions from direct likelihood-maximizing approaches on
synthetic data. We would like to point out some nice additional
properties of our formulation. First, as the dimensionality of the
input data increases, so does the probability that the random
direction is approximately orthogonal to the true
relationship. Second, due to the presence of the parameter $\alpha$ in
the loss, one may easily tune the focus on performance versus
privacy. Third, the factor of dataset reduction in order to construct
a core-set depends on the magnitude of the unexplained variance---for
very low unexplained variance, one needs many samples just to find a
few unrepresentative examples. In more difficult, real-world problems
though, this should actually be even easier. One potential
vulnerability is that if an adversary has good priors on the
distribution of input features, they may be able to gain information
through the difference in feature distribution in the core-set.  In the
future, we would like to address these problems and experiment with
regressions on real-world data.

Additionally, it should be noted that the CIFAR-100 task we chose is actually an especially difficult problem for this approach. First, the size of the dataset is only several times larger than the amount of data required to reasonably consistently train a CNN from scratch, which limits choice of unrepresentative samples. Second, the tasks of predicting class versus super-class are very similar, which makes it surprising that it is even possible to find core-sets which are significantly more predictive on one task than the other.  It will be interesting to try these ideas on other datasets such as the ImageNet.

\section{Conclusions}

In this work we have presented simple and practical methods for constructing fair core-sets which better preserve privacy. We demonstrate their utility on both synthetic and real-world datasets, with both linear and neural models. We hope that this work will pave the way for smaller and more private datasets in the future.

\subsubsection*{Acknowledgments}

We'd like to thank Tushar Chandra for productive conversation related to this work, and Asher Spector for advice, comments, and revisions on this paper.

\nocite{*}

\bibliographystyle{neurips_2019}
\bibliography{neurips_2019}

\end{document}